\documentclass[conference]{IEEEtran}
\IEEEoverridecommandlockouts

\usepackage{cite}
\usepackage{amsmath,amssymb,amsfonts}
\usepackage{algorithmic}
\usepackage{graphicx}
\usepackage{textcomp}
\usepackage{xcolor}

\usepackage{booktabs}

\def\BibTeX{{\rm B\kern-.05em{\sc i\kern-.025em b}\kern-.08em
    T\kern-.1667em\lower.7ex\hbox{E}\kern-.125emX}}
    
\begin{document}
\title{Brick-Diffusion: Generating Long Videos with Brick-to-Wall Denoising}

\makeatletter
\newcommand{\newlineauthors}{%
  \end{@IEEEauthorhalign}\hfill\mbox{}\par
  \mbox{}\hfill\begin{@IEEEauthorhalign}
}
\makeatother

\author{\IEEEauthorblockN{1\textsuperscript{st} Yunlong Yuan}
\IEEEauthorblockA{\textit{School of Data Science} \\
\textit{Fudan University}\\
Shanghai, China \\
ylyuan22@m.fudan.edu.cn}
\and
\IEEEauthorblockN{2\textsuperscript{nd} Yuanfan Guo}
\IEEEauthorblockA{\textit{Noah's Ark Lab} \\
\textit{Huawei}\\
Shanghai, China \\
guoyuanfan1@huawei.com}
\and
\IEEEauthorblockN{3\textsuperscript{rd} Chunwei Wang}
\IEEEauthorblockA{\textit{Noah's Ark Lab} \\
\textit{Huawei}\\
Shanghai, China \\
wangchunwei5@huawei.com}
\newlineauthors
\IEEEauthorblockN{4\textsuperscript{th} Hang Xu}
\IEEEauthorblockA{\textit{Noah's Ark Lab} \\
\textit{Huawei}\\
Shanghai, China \\
chromexbjxh@gmail.com}
\and
\IEEEauthorblockN{5\textsuperscript{th} Li Zhang\textsuperscript{\dag}\thanks{$\dag$ Li Zhang is the corresponding author.}}
\IEEEauthorblockA{\textit{School of Data Science} \\
\textit{Fudan University}\\
Shanghai, China \\
lizhangfd@fudan.edu.cn}
}

\maketitle

\begin{abstract}
Recent advances in diffusion models have greatly improved text-driven video generation. However, training models for long video generation demands significant computational power and extensive data, leading most video diffusion models to be limited to a small number of frames.
Existing training-free methods that attempt to generate long videos using pre-trained short video diffusion models often struggle with issues such as insufficient motion dynamics and degraded video fidelity.
In this paper, we present Brick-Diffusion, a novel, training-free approach capable of generating long videos of arbitrary length. 
Our method introduces a brick-to-wall denoising strategy, where the latent is denoised in segments, with a stride applied in subsequent iterations.
This process mimics the construction of a staggered brick wall, where each brick represents a denoised segment, enabling communication between frames and improving overall video quality.
Through quantitative and qualitative evaluations, we demonstrate that Brick-Diffusion outperforms existing baseline methods in generating high-fidelity videos.
\end{abstract}

\begin{IEEEkeywords}
Diffusion models, long video generation, brick-to-wall denoising, training-free.
\end{IEEEkeywords}

\section{Introduction}
Long video generation, defined as the creation of videos comprising at least hundreds of frames, is a research area of significant importance due to its potential applications in content creation, media, and real-world simulation.
Recent advancements in diffusion models~\cite{Diffusion,DDPM,Score-Based}, particularly in the image domain~\cite{StableDiffusion,SDXL,GLIDE,PixArt,DiT,Cascaded}, have laid the groundwork for the development of video diffusion models~\cite{VDM, MCVD, Align,VideoCrafter2,AnimateDiff,SVD,ModelScope, Latte, Make-A-Video, MagicVideo, PYoCo, LVDM}.
However, the majority of existing text-to-video diffusion models~\cite{VDM, VideoCrafter2,AnimateDiff,SVD, ModelScope, MagicVideo} are primarily focused on generating short videos, typically around 2 seconds in length with 16 frames.
Nevertheless, the short video diffusion models hold the potential to generate long videos. 
The area of long video generation using short video diffusion models is not yet fully explored and presents a promising avenue for further research.
\begin{figure}[t]
\begin{center}
    \includegraphics[width=0.99\linewidth]{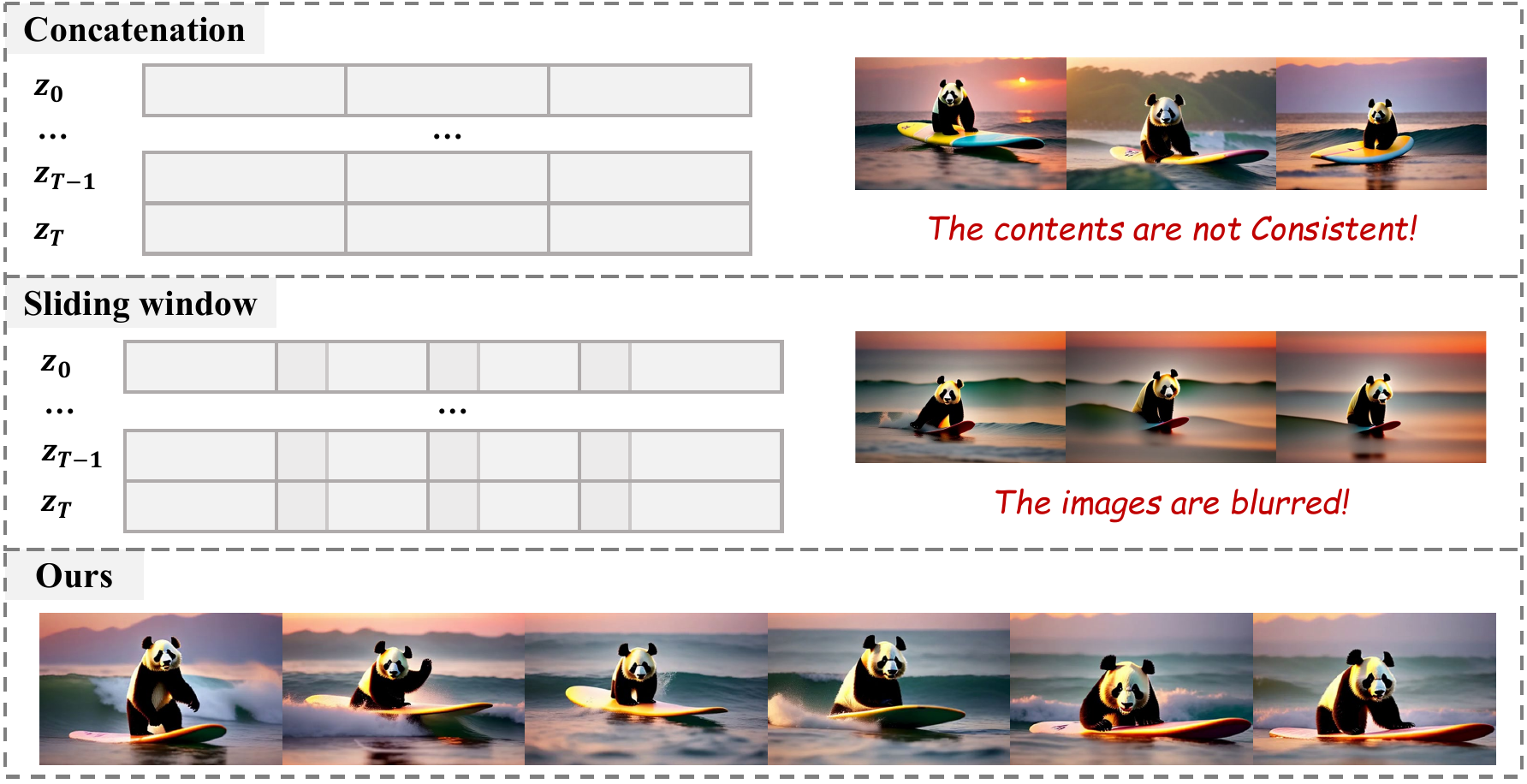}
\end{center}
\caption{Comparisons between different methods for long video generation.  
(a) \textbf{Concatenation}: denoises segments individually and concatenates them.
(b) \textbf{Sliding window}: denoises segments in a sliding window approach.
(c) \textbf{Ours}: uses the brick-to-wall denoising, generating videos with high fidelity.
}
\label{fig:intro}
\end{figure}

\begin{figure*}[t]
\begin{center}
    \includegraphics[width=0.95\linewidth]{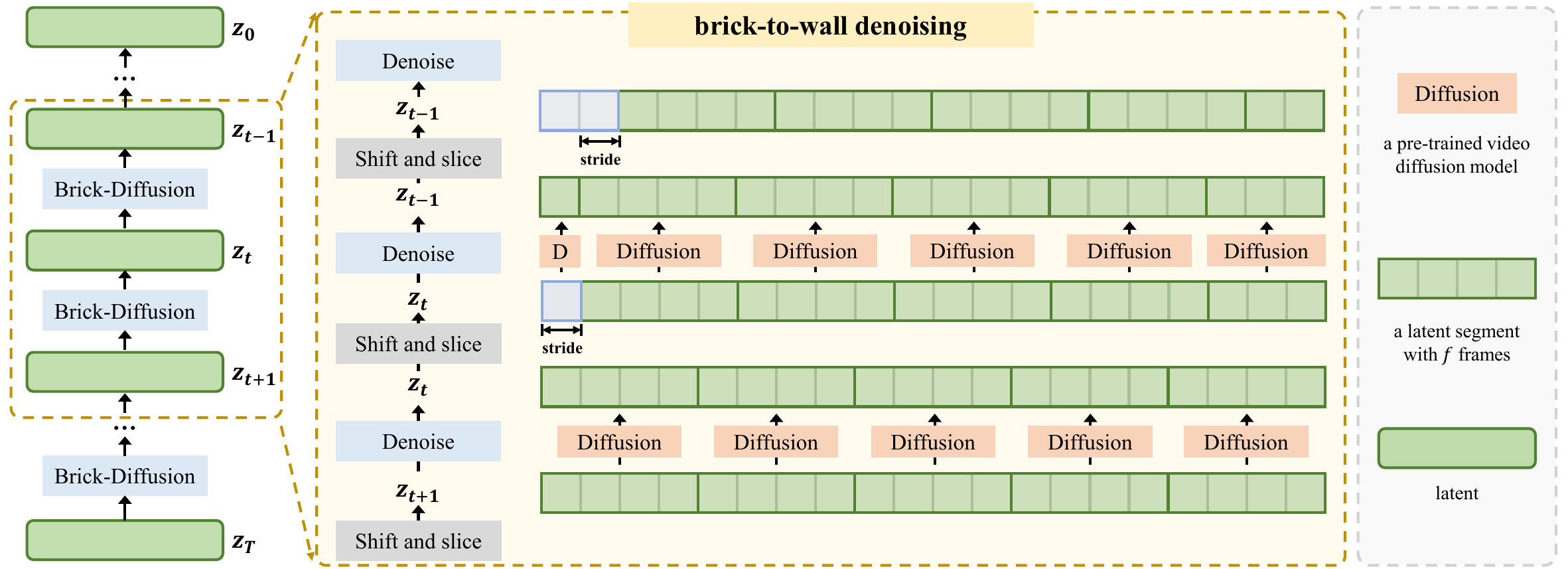}
\end{center}
\caption{The framework of Brick-Diffusion. For each denoising step, we slice the latent into segments and denoise them individually using a diffusion model. In the subsequent step, we apply a stride to shift and re-slice the latent into new segments. This process is repeated until we obtain the final clean latent.}
\label{fig:pipeline}
\end{figure*}
Generating long videos presents substantial challenges due to inherent temporal complexity, resource constraints, and the need to maintain content consistency throughout the video. 
In response to the growing attention to this task, various methods have been proposed, including GAN-based methods~\cite{StyleGAN-v, LongVideoGAN, Time-Agnostic-VQGAN} and diffusion-based methods~\cite{FlexibleDiffusion,MCVD,LVDM, NUWA-XL,Phenaki,SEINE}.
However, these approaches require extensive training on large video datasets. 
Therefore, some recent works~\cite{Gen-L-Video, FreeNoise, FIFO} have explored training-free methods that focus on the inference stage of diffusion models to generate long videos.
A straightforward approach is concatenation, where short clips are generated using diffusion models and then stitched together, as shown in Fig.~\ref{fig:intro}. 
However, this method fails to ensure consistency between clips, resulting in noticeable flickering at the junctions. 
Gen-L-Video~\cite{Gen-L-Video} addresses this flickering issue by applying a sliding-window approach to denoise the latent and merge overlapping sub-segments through averaging, as depicted in Fig.~\ref{fig:intro}. 
Unfortunately, this averaging process removes necessary noise from the latent, altering its distribution and resulting in blurred videos with reduced fidelity.
FreeNoise~\cite{FreeNoise} introduces window-based attention fusion during inference, but this method demands significant memory resources, especially when computing attention weights in the temporal attention layers across many frames.
Moreover, the generated videos tend to appear overly smooth and lack motion dynamics.
FIFO-Diffusion~\cite{FIFO} proposes a diagonal-denoising technique, generating videos frame by frame using a queue structure. 
However, this approach suffers from a training-inference gap, as the noise level differs across frames in diagonal denoising, whereas, during training, all frames in the latent are corrupted by the same level of noise. 
This gap negatively impacts the quality of the generated videos.

To address these limitations, we propose a novel method called \textbf{Brick-Diffusion} for generating long videos. 
The core of our approach is the brick-to-wall denoising process. 
Specifically, we slice the latent into short segments and denoise each segment individually using a short video diffusion model. In the subsequent step, we apply a stride to shift and re-slice the latent. 
This process resembles constructing a wall with staggered bricks, where each layer of bricks is offset relative to the previous one.  
The brick-to-wall denoising facilitates effective communication between different segments, resulting in consistent and high-quality long videos, as illustrated in Fig.~\ref{fig:intro}. Besides, unlike FreeNoise~\cite{FreeNoise} and FIFO-Diffusion~\cite{FIFO}, our method can be easily implemented in a parallized manner.

In summary, our contributions are as follows:
\begin{itemize}
    \item We propose \textbf{Brick-Diffusion}, a novel framework that can generate long videos of any length using pre-trained video diffusion models without the need for fine-tuning and can be easily implemented in a parallelized manner.
    \item We design the brick-to-wall denoising technique, which resembles constructing a staggered brick wall, producing videos with high quality and fidelity.
    \item Qualitative and quantitative experiments demonstrate the effectiveness and superiority of our method.
\end{itemize}

\section{Brick-Diffusion}
The task of generating long videos using short video diffusion models can be formulated as follows: Given a short diffusion model $\epsilon_\theta$ designed to generate $f$ frames, the goal is to generate a video $v$ consisting of $F$ frames ($F>f$, e.g., $F=128$ and $f=16$).

\subsection{Preliminary}
Diffusion models~\cite{Diffusion, DDPM,Score-Based} are probabilistic generative models that are trained to learn a data distribution by the gradual denoising of a variable sampled from a Gaussian distribution.
A general video diffusion model consists of two components: an auto-encoder $\mathrm{Enc}(\cdot)$ and $\mathrm{Dec}(\cdot)$, and a noise prediction network $\epsilon_\theta(\cdot)$. For the Gaussian noise $\epsilon$, the diffusion timestep $t$, and the text condition $c$, the diffusion model is trained to minimize the following $l_2$ loss:
\begin{equation}
    \min_\theta \mathrm{E}_{z_0,\epsilon\sim\mathcal{N}(0,I),t\sim \mathrm{Uniform}(1, T)} ||\epsilon - \epsilon_\theta(z_t, t, c)||_2^2.
\end{equation}
Given a diffusion model $\epsilon_\theta$, timestep $t$, and the condition $c$, we can predict the noise, and denoise the latent $z_t$ as follows:
\begin{equation}
    z_{t-1} = \Phi(z_t, t, c;\epsilon_\theta),
\end{equation}
where $\Phi(\cdot)$ represents any diffusion model sampler~\cite{DDIM,DPM-Solver,EDM,PNDM}. 
DDIM~\cite{DDIM} is widely used and serves as the default sampler for many diffusion models. We select the DDIM sampler in our method.

\subsection{Brick-to-wall denoising}
As illustrated in Fig.~\ref{fig:pipeline}, we present the approach of Brick-Diffusion. 
The entire process follows a typical diffusion sampling procedure, but to handle latents with a large number of frames, we employ the brick-to-wall denoising technique: we slice the latent every $f$ frames, dividing the entire latent into segments, each with a length of $f$ frames, which matches the default input size of the pre-trained diffusion model. 
Then we denoise each segment individually.
For subsequent denoising steps, we first shift the slices with a stride and re-slice the latent, and then denoise the latent.
This process is akin to constructing a wall with bricks, where each layer of bricks is staggered relative to the previous one.
The two key procedures of brick-to-wall denoising are: (1) slice the latent into segments; and (2) denoise each segment.

\paragraph{Slice the latent into segments}
Before denoising, we need to slice the long latent into segments.
Since the nature of shifting operation, we only need to calculate the offset to determine the positions of each segment. 
Specifically, at the beginning, the offset is set to $0$. At the timestep of $t$, the offset is calculated as follows:
\begin{equation}
    \mathrm{offset}_t = \mathrm{stride} \times (T - t) - \lfloor\frac{\mathrm{stride} \times (T - t)}{f}\rfloor \times f.
\end{equation}
From this, we infer the following equation:
\begin{equation}
    \mathrm{offset}_t \equiv \mathrm{offset}_{t+1} + \mathrm{stride} \pmod{f}.
\end{equation}
Therefore, the offset at timestep $t$ will be the previous offset shifted by a stride, followed by a modulo operation with $f$, ensuring that the offset cycles within the range of frame lengths.
At timestep $t$, we shift and re-slice the latent $z_t$ according to $\mathrm{offset}_t$. 
If the offset is non-zero, the first segment will be $z_t^{0:\mathrm{offset}_t}$, representing the first $\mathrm{offset}_t$ frames of the latent $z_t$.
For the $i$-the segment ($i>1$), the start index and end index are determined as follows:
\begin{equation}
    \mathrm{start} = (i-1)\times f + \mathrm{offset}_t, \quad \mathrm{end} = i\times f + \mathrm{offset}_t.
\end{equation}
Therefore, the $i$-th segment is $z_t^{\mathrm{start}:\mathrm{end}}$.

\paragraph{Denoise each segment}
We denoise segment $z_t^{\mathrm{start}:\mathrm{end}}$ like the general sampling process of diffusion models: 
\begin{equation}
    z_{t-1}^{\mathrm{start}:\mathrm{end}} = \Phi(z_{t}^{\mathrm{start}:\mathrm{end}}, t, c; \epsilon_\theta). 
\end{equation}
However, slicing the latent may result in the first and last segments being shorter than the common segment length $f$.
If $\mathrm{offset}_t$ is non-zero, the first segment will contain fewer than $f$ frames. To address this, we extend the segment to a common size and denoise it as follows: 
\begin{equation}
    z_{temp}^{0:f} = \Phi(z_{t}^{0:f}, t, c; \epsilon_\theta). 
\end{equation}
Next, we update only the first $\mathrm{offset}_t$ frames of the extended segment and discard the remainder:
\begin{equation}
    z_{t-1}^{0:\mathrm{offset}_t} = z_{temp}^{0:\mathrm{offset}_t}.
\end{equation}
For the last segment, if it is shorter than $f$, we apply the same procedure. After denoising all segments of $z_t$, we concatenate them to form the latent $z_{t-1}$. 
Since each segment is processed independently, the brick-to-wall denoising can be efficiently implemented in a parallelized manner.

The brick-to-wall denoising repeats the two procedures until we get the clean latent $z_0$. Then passing $z_0$ through a decoder $\mathrm{Dec}(\cdot)$, we get the generated long video $v =\mathrm{Dec}(z_0)$.

In practice, to eliminate the influence of denoising the shorter segments, we pad the initial latent $z_t$ to $F+2f$ frames and take the middle $F$ frames of the video $v$ as the final result.

Shifting with a non-zero stride during latent slicing is crucial for enabling communication between different video frames.
Without this shifting, the method reduces to simple concatenation.
Shifting ensures that the information between adjacent segments is fused and interacted with during the next denoising step. As the denoising progresses, each frame of the latent will have sufficient communication with a broad range of other frames, making the final video consistent.
Moreover, unlike the window-based attention used in FreeNoise~\cite{FreeNoise} and the frame-by-frame generation method employed by FIFO-Diffusion~\cite{FIFO}, our method is easily parallelizable, enabling more efficient processing.

\begin{figure*}[t]
\centerline{\includegraphics[width=1.0\linewidth]{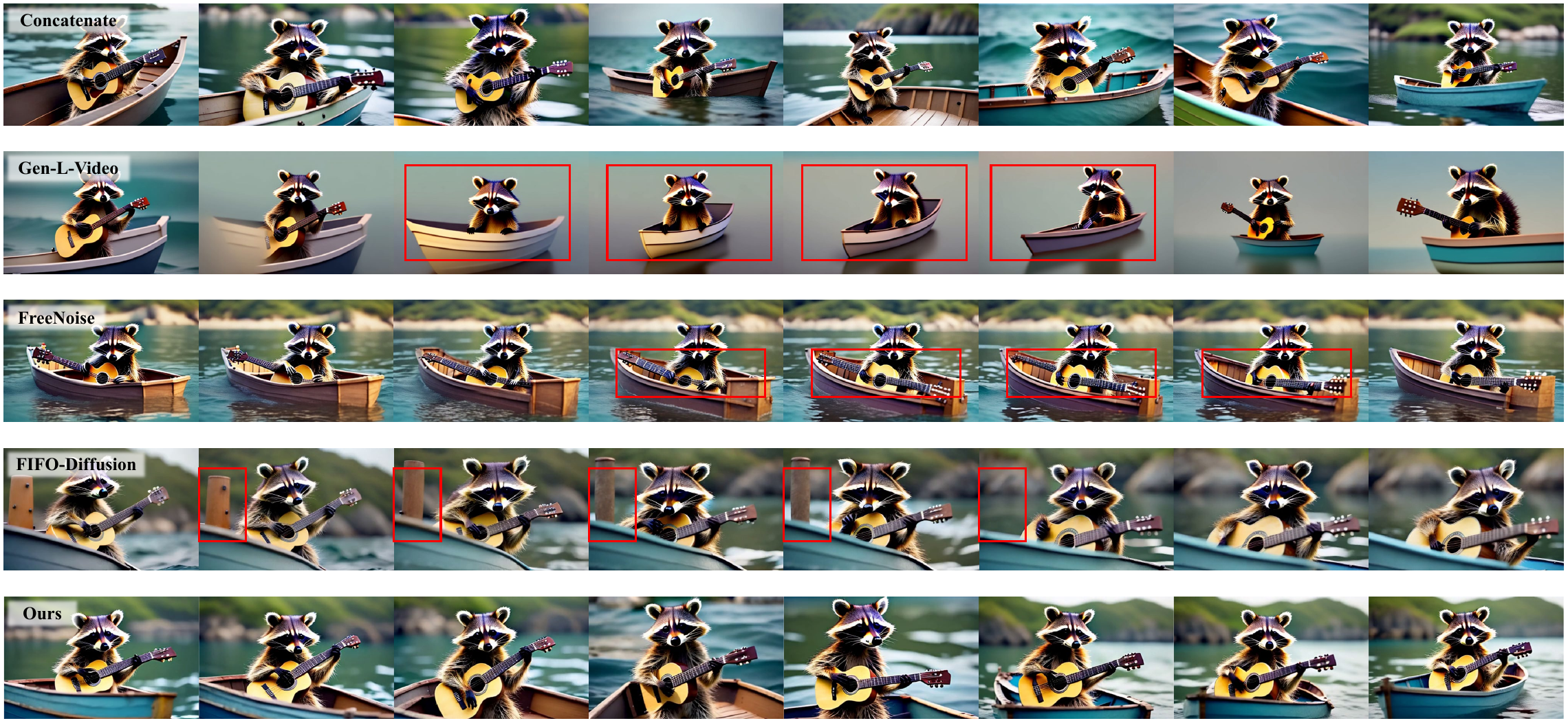}}
\caption{Qualitative results of each method. The text prompt is ``a cute raccoon playing guitar in a boat on the ocean.'' The method of directly concatenating clips results in dramatic content changes. For the other baseline methods, we use red boxes to highlight the issues present in the generated video frames.}
\label{fig:demo}
\end{figure*}

\section{Experiments}
\subsection{Benchmarks and Evaluation Metrics}
We use VBench~\cite{VBench} as the benchmark for our experiments and report the following metrics: subject consistency~\cite{DINOv2}, dynamic degree~\cite{RAFT}, aesthetic quality~\cite{Aesthetic}, and overall video-text consistency~\cite{ViCLIP}.
To facilitate meaningful comparisons, we normalize these metric scores to a range of 0 to 100 based on the empirical maximum and minimum values reported in VBench~\cite{VBench}.
Subject consistency measures the consistency of subjects within a video, while dynamic degree assesses the level of motion dynamics.  
Aesthetic quality evaluates the artistic and visual appeal, considering factors such as layout, color richness, and harmony.
Overall video-text consistency calculates the similarity between video features and text features, reflecting the semantic alignment.

\subsection{Baselines}
To evaluate the effectiveness and generalization of our proposed method, we select four methods as baselines for our experiments. 
These methods includes direct concatenation, Gen-L-Video~\cite{Gen-L-Video}, FreeNoise~\cite{FreeNoise}, and FIFO-Diffusion~\cite{FIFO}.
All these four methods are training-free.

\subsection{Implementation Details} 
All baseline methods, as well as our method, are tasked with generating long videos consisting of 128 frames during the inference stage, 
using the open-source text-to-video diffusion model VideoCrafter2~\cite{VideoCrafter2}, which is designated to generate 16-frame videos at a resolution of $320 \times 512$.
By default, We employ the DDIM sampler~\cite{DDIM}. 
Each method is required to generate 825 videos for evaluation using the prompts from VBench~\cite{VBench}. For our method, the stride $s$ is set to 1.

\subsection{Results}
\paragraph{Quantitative results}
In Table~\ref{tab:main_table}, we present the quantitative results of our proposed method alongside other baseline methods.
Our method achieves the highest scores in dynamic degree, aesthetic quality, and overall video-text consistency. While FreeNoise~\cite{FreeNoise} performs the best on subject consistency, it does so at the expense of dynamic degree, for which it receives the lowest score of 48.02.
Considering all four metrics, we calculate the average score, where our method achieves 81.70, surpassing the second-best, FIFO-Diffusion~\cite{FIFO}, which scores 79.12, indicating the superiority of our method.

\begin{table}[tbh]
\caption{Quantitative comparison between our method and baselines.}
\begin{center}
    \begin{tabular}{c|cccc|c}
    \toprule
    Methods       & Sub $\uparrow$ & Dyn $\uparrow$ & Aes $\uparrow$ & Text $\uparrow$ & Avg $\uparrow$ \\
    \midrule
    Concatenate   &     91.56      &      57.36     &     59.80      &   80.08         &     72.20      \\
    Gen-L-Video   &     86.62      &      49.48     &     52.69      &   64.34         &     63.28      \\
    FreeNoise     & \textbf{95.09} &      48.02     &     59.75      &   78.57         &     70.36      \\
    FIFO-Diffusion&     93.91      &      83.36     &     59.60      &   79.59         &     79.12      \\
    \midrule     
    Ours          &     93.00      &  \textbf{92.52}& \textbf{59.84} & \textbf{81.43}  & \textbf{81.70} \\ 
    \bottomrule
    \end{tabular}
    \label{tab:main_table}
\end{center}
\end{table}

\paragraph{Qualitative results} 
As shown in Fig~\ref{fig:demo}, directly concatenating short clips does not produce a consistent long video, leading to noticeable content changes between frames. 
It can be expected, as each short clip does not have information communication with other clips.
Gen-L-Video~\cite{Gen-L-Video} averages the overlapping areas, which adversely affects the latent, resulting in a loss of fidelity in those frames. 
The details of the background are missing, making the frame blurred.
Besides, the guitar is missing in the middle of four frames, causing inconsistency between the text prompt and the generated video.
The output generated by FreeNoise~\cite{FreeNoise} lacks dynamic variation, with minimal scene changes across frames. However, the guitar in the red box has unrealistic deformations.
In the video generated by FIFO-Diffusion~\cite{FIFO}, the object on the left side of the frames changes and eventually disappears. 
In contrast, our method produces high-fidelity long videos with high dynamics, outperforming all other methods.

\paragraph{Ablation Study}
We explore the effects of different stride values. When the stride $s$ equals $0$, our method, Brick-Diffusion, essentially degrades into directly concatenating short clips. 
As shown in Table~\ref{tab:stride}, the performance with a non-zero stride is significantly better than with direct concatenation ($s=0$). 
The results for $s=1$, $s=3$, $s=5$, $s=7$, and $s=9$ vary across the four metrics, but overall, the average scores are relatively close. 
We select a stride of $s=1$ for our final results, as it achieves the highest average score.

\begin{table}[tbh]
\caption{Comparison of the influence of the stride.}
\begin{center}
    \begin{tabular}{c|cccc|c}
    \toprule
    Stride $s$  & Sub $\uparrow$ & Dyn $\uparrow$ & Aes $\uparrow$ & Text $\uparrow$ & Avg $\uparrow$ \\
    \midrule
    $s=0$       &     91.56      &      57.36     &     59.80      &   80.08         &     72.20      \\
    \midrule
    $s=1$       &     93.00      &      92.52     &     59.84      &   81.43         &  \textbf{81.70}\\
    $s=3$       &     93.04      &      90.01     &     60.01      &   81.15         &     81.06      \\
    $s=5$       &     93.10      &      90.03     &     60.03      &   81.26         &     81.11      \\
    $s=7$       &     93.25      &      89.92     &     60.03      &   81.24         &     81.11      \\
    $s=9$       &     93.33      &      89.83     &     59.99      &   81.24         &     81.10      \\ 
    \bottomrule
    \end{tabular}
    \label{tab:stride}
\end{center}
\end{table}

\section{Conclusion}
We propose Brick-Diffusion, a novel approach for long video generation that leverages a pre-trained video diffusion model with brick-to-wall denoising strategy. 
Our method enables communications between segments, ensuring consistency and high fidelity in generated videos.
The experimental results demonstrate that Brick-Diffusion outperforms existing baseline methods, highlighting its effectiveness.

\section*{Acknowledgment}
This work was supported in part by National Natural Science Foundation of China (Grant No. 62376060) and
Natural Science Foundation of Shanghai (Grant No. 22ZR1407500).


\begin{thebibliography}{00}
\bibitem{Diffusion} J. Sohl-Dickstein, E. Weiss, N. Maheswaranathan, and S. Ganguli, ``Deep unsupervised learning using nonequilibrium thermodynamics,'' in \textit{ICML}, 2015.
\bibitem{DDPM} J. Ho, A. Jain, and P. Abbeel, ``Denoising diffusion probabilistic models,'' in \textit{NeurIPS}, 2020.
\bibitem{Score-Based} Y. Song, J. Sohl-Dickstein, D. P. Kingma, and et al., ``Score-based generative modeling through stochastic differential equations,'' in \textit{ICLR}, 2021.

\bibitem{StableDiffusion} R. Rombach, A. Blattmann, D. Lorenz, P. Esser, and B. Ommer, ``High-resolution image synthesis with latent diffusion models,'' in \textit{CVPR}, 2022.
\bibitem{GLIDE} A. Q. Nichol, P. Dhariwal, A. Ramesh, and et al., ``GLIDE: towards photorealistic image generation and editing with text-guided diffusion models,'' in \textit{ICML}, 2022.
\bibitem{Cascaded} J. Ho, C. Saharia, W. Chan, D. J. Fleet, M. Norouzi, and T. Salimans, ``Cascaded Diffusion Models for High Fidelity Image Generation,'' \textit{JMLR}, 2022.
\bibitem{DiT} W. Peebles and S. Xie, ``Scalable diffusion models with transformers,''
in \textit{ICCV}, 2023.
\bibitem{SDXL} D. Podell, Z. English, K. Lacey, and et al., ``SDXL: improving latent diffusion models for high-resolution image synthesis,'' in \textit{ICLR}, 2024.
\bibitem{PixArt} J. Chen, J. YU, C. GE, and et al., ``PixArt-$\alpha$: fast training of diffusion transformer for photorealistic text-to-image synthesis,'' in \textit{ICLR}, 2024.

\bibitem{VideoCrafter2} H. Chen, Y. Zhang, X. Cun, and et al., ``Videocrafter2: overcoming data limitations for high-quality video diffusion models,'' in \textit{CVPR}, 2024.
\bibitem{AnimateDiff} Y. Guo, C. Yang, A. Rao, and et al., ``AnimateDiff: animate your personalized text-to-image diffusion models without specific tuning,'' in \textit{ICLR}, 2024.
\bibitem{SVD} A. Blattmann, T. Dockhorn, S. Kulal, and et al., ``Stable video diffusion: scaling latent video diffusion models to large datasets,'' 2023, arXiv:2311.15127. [Online]. Available: https://arxiv.org/abs/2311.15127

\bibitem{ModelScope} J. Wang, H. Yuan, D. Chen, and et al., ``ModelScope text-to-video technical report,'' 2023, arXiv:2308.06571. [Online]. Available: https://arxiv.org/abs/2308.06571
\bibitem{MagicVideo} D. Zhou, W. Wang, H. Yan, W. Lv, Y. Zhu, and J. Feng, ``MagicVideo: efficient video generation with latent diffusion models,'' 2023, arXiv:2211.11018. [Online]. Available: https://arxiv.org/abs/2211.11018

\bibitem{VDM} J. Ho, T. Salimans, A. Gritsenko, W. Chan, M. Norouzi, and D. J. Fleet, ``Video diffusion models,'' in \textit{NeurIPS}, 2022.
\bibitem{Align}
A. Blattmann, R. Rombach, H. Ling, and et al., ``Align your latents: high-resolution video synthesis with latent diffusion models,'' in \textit{CVPR}, 2023.
\bibitem{MCVD} V. Voleti, A. Jolicoeur-Martineau, and C. Pal, ``MCVD: masked conditional video diffusion for prediction, generation, and interpolation,'' in \textit{NeurIPS}, 2022.
\bibitem{LVDM} Y. He, T. Yang, Y. Zhang, Y. Shan, and Q. Chen, ``Latent Video Diffusion Models for High-Fidelity Long Video Generation,'' 2023, arXiv:2211.13221. [Online]. Available: https://arxiv.org/abs/2211.13221
\bibitem{Latte} X. Ma, Y. Wang, G. Jia, and et al., ``Latte: latent diffusion transformer for video generation,'' 2024, arXiv:2401.03048. [Online]. Available: https://arxiv.org/abs/2401.03048
\bibitem{Make-A-Video} U. Singer, A. Polyak, T. Hayes, and et al., ``Make-A-Video: text-to-video generation without text-video data,'' in \textit{ICLR}, 2023.
\bibitem{PYoCo} S. Ge, S. Nah, G. Liu, and et al., ``Preserve your own correlation: a noise prior for video diffusion models,'' in \textit{ICCV}, 2023.

\bibitem{StyleGAN-v} I. Skorokhodov, S. Tulyakov, and M. Elhoseiny, ``StyleGAN-V: a continuous video generator with the price, image quality and perks of StyleGAN2,'' in \textit{CVPR}, 2022.
\bibitem{LongVideoGAN}
T. Brooks, J. Hellsten, M. Aittala, and et al., ``Generating long videos of dynamic scenes,'' in \textit{NeurIPS}, 2022.
\bibitem{Time-Agnostic-VQGAN} S. Ge, T. Hayes, H. Yang, and et al., ``Long video generation with time-agnostic VQGAN and time-sensitive transformer,'' in \textit{ECCV}, 2022.

\bibitem{FlexibleDiffusion} W. Harvey, S. Naderiparizi, V. Masrani, C. Weilbach, and F. Wood, ``Flexible diffusion modeling of long videos,'' in \textit{NeurIPS}, 2022.
\bibitem{NUWA-XL} S. Yin, C. Wu, H. Yang, and et al., ``NUWA-XL: diffusion over diffusion for extremely long video generation,'' 2023, arXiv:2303.12346. [Online]. Available: https://arxiv.org/abs/2303.12346
\bibitem{Phenaki} R. Villegas, M. Babaeizadeh, P. J. Kindermans, and et al., ``Phenaki: variable length video generation from open domain textual descriptions,'' in \textit{ICLR}, 2023.
\bibitem{SEINE} X. Chen, Y. Wang, L. Zhang, and et al., ``SEINE: short-to-long video diffusion model for generative transition and prediction,'' in \textit{ICLR}, 2024.

\bibitem{Gen-L-Video} F. Wang, W. Chen, G. Song, and et al., ``Gen-L-Video: multi-text to long video generation via temporal co-denoising,'' 2023, arXiv.2305.18264. [Online]. Available: https://arxiv.org/abs/2305.18264
\bibitem{FreeNoise} H. Qiu, M. Xia, Y. Zhang, ane et al., ``FreeNoise: tuning-free longer video diffusion via noise rescheduling,'' in \textit{ICLR}, 2024.
\bibitem{FIFO}
J. Kim, J. Kang, J. Choi, and B. Han, ``FIFO-Diffusion: generating infinite videos from text without training,'' 2024, arXiv:2405.11473. [Online]. Available: https://arxiv.org/abs/2405.11473

\bibitem{DDIM} J. Song, C. Meng, and S. Ermon, ``Denoising diffusion implicit models,'' in \textit{ICLR}, 2021.
\bibitem{DPM-Solver} C. Lu, Y. Zhou, F. Bao, J. Chen, C. LI, and J. Zhu, ``DPM-Solver: a fast ODE solver for diffusion probabilistic model sampling in around 10 steps,'' in \textit{NeurIPS}, 2022.
\bibitem{EDM} T. Karras, M. Aittala, T. Aila, and S. Laine, ``Elucidating the design space of diffusion-based generative models,'' in \textit{NeurIPS}, 2022.
\bibitem{PNDM} L. Liu, Y. Ren, Z. Lin, and Z. Zhao, ``Pseudo numerical methods for diffusion models on manifolds,'' in \textit{ICLR}, 2022.

\bibitem{VBench} Z. Huang, Y. He, J. Yu, and et al., ``VBench: comprehensive benchmark suite for video generative models,'' in \textit{CVPR}, 2024.
\bibitem{DINOv2} M. Oquab, T. Darcet, T. Moutakanni, and et al., ``DINOv2: learning robust visual features without supervision,'' \textit{TMLR}, 2024.
\bibitem{RAFT} Z. Teed and J. Deng, ``RAFT: recurrent all-pairs field transforms for optical flow,'' in \textit{ECCV}, 2020.
\bibitem{Aesthetic} LAION-AI, ``Aesthetic predictor,'' 2022, gitHub repository. [Online]. Available: https://github.com/LAION-AI/aesthetic-predictor
\bibitem{ViCLIP} Y. Wang, Y. He, Y. Li, and et al., ``InternVid: a large-scale video-text dataset for multimodal understanding and generation,'' in \textit{ICLR}, 2024.



\end{thebibliography}
\end{document}